\documentclass[8pt]{article}
\usepackage[utf8]{inputenc}
\usepackage[a4paper, portrait, margin=1in]{geometry}
\usepackage{comment}
\usepackage{float}
\usepackage[dvipsnames]{xcolor}
\usepackage{bbm}
\usepackage{makecell}
\usepackage{multirow}
\usepackage{multicol}
\usepackage{subcaption}
\usepackage{lscape}
\usepackage{tikz}
\usetikzlibrary{arrows,decorations.pathmorphing,backgrounds,positioning,fit,petri}
\usetikzlibrary{bayesnet}
\usetikzlibrary{mindmap,trees}
\definecolor{coolred}{rgb}{0.0, 0.2, 0.8}

\usepackage{helvet}
\usepackage[backend=biber, maxcitenames = 2, mincitenames=1,
citestyle=authoryear, giveninits=true, isbn=false, doi=true, url = true,
maxbibnames=99, sorting=nyt, style=authoryear,
uniquelist=false,uniquename=init,sortcites]{biblatex}

\bibliography{BayesianRCM}
\renewbibmacro*{in:}{%
  \iffieldundef{journaltitle}
    {}
    {\printtext{\bibstring{in}\intitlepunct}}}

\newcommand*{\tabindent}{ \hspace{3mm}}

\newcommand{\N}{\mathcal{N}}
\newcommand{\E}{\mathbb{E}}
\newcommand{\bs}{\boldsymbol}
\newcommand{\bx}{\textbf{x}}
\newcommand{\bX}{\textbf{X}}

\newcommand{\by}{\textbf{y}}

\newcommand{\bz}{\textbf{z}}

\definecolor{mygray}{gray}{0.6}

\usepackage{authblk}
\usepackage{amsmath,amssymb}
\usepackage{bm}

\usepackage{accents}
\newcommand{\dbtilde}[1]{\accentset{\approx}{#1}}

\title{Context-aware Bayesian Mixed Multinomial Logit Model}
\author[$\ast$,$\dagger$]{Miros\l{}awa \L{}ukawska}
\author[$\dagger$]{Anders Fjendbo Jensen}
\author[$\dagger$]{Filipe Rodrigues}
\affil[$\ast$]{Corresponding author (mirlu@dtu.dk)}
\affil[$\dagger$]{Technical University of Denmark, Department of Technology, Management and Economics, Bygningstorvet 116b, 2800 Kgs. Lyngby}
\date{}
\usepackage{graphicx}
\usepackage[colorlinks=true, allcolors=blue]{hyperref}
\usepackage{cleveref}

\usepackage{color}

\def\tg{\color{gray}}

\usepackage{appendix}

\usepackage{float}

\begin{document}
\maketitle
\noindent

\vspace{2em}
\textbf{Declarations of interest:} none

\textbf{Highlights}
\begin{itemize}
    \item We propose an approach to model context-dependent intra-respondent heterogeneity.
    \item Contextual information is mapped to additive shifts on preference parameters.
    \item The model allows non-linear interactions between continuous and discrete variables.
    \item Analyses of different context scenarios are possible without model reestimation.
    \item We estimate a C-MMNL bicycle route choice model on 119,448 trips in 47 minutes.
\end{itemize}

\section*{Abstract}
The mixed multinomial logit model assumes constant preference parameters of a decision-maker throughout different choice situations, which may be considered too strong for certain choice modelling applications. This paper proposes an effective approach to model context-dependent intra-respondent heterogeneity, thereby introducing the concept of the Context-aware Bayesian mixed multinomial logit model, where a neural network maps contextual information to interpretable shifts in the preference parameters of each individual in each choice occasion. The proposed model offers several key advantages. First, it supports both continuous and discrete variables, as well as complex non-linear interactions between both types of variables. Secondly, each context specification is considered jointly as a whole by the neural network rather than each variable being considered independently. Finally, since the neural network parameters are shared across all decision-makers, it can leverage information from other decision-makers to infer the effect of a particular context on a particular decision-maker. Even though the context-aware Bayesian mixed multinomial logit model allows for flexible interactions between attributes, the increase in computational complexity is minor, compared to the mixed multinomial logit model. We illustrate the concept and interpretation of the proposed model in a simulation study. We furthermore present a real-world case study from the travel behaviour domain - a bicycle route choice model, based on a large-scale, crowdsourced dataset of GPS trajectories including 119,448 trips made by 8,555 cyclists.\\

\textbf{Keywords:} choice context, bayesian modelling, neural networks, bicycle route choice, big data

\section{Introduction}
%Introduction State the objectives of the work and provide an adequate background, avoiding a detailed literature survey or a summary of the results.

A long-standing major concern in discrete choice modelling is heterogeneity in the decision-making process. To overcome this, the mixed multinomial logit (MMNL) model \parencite{revelt1998mixed, mcfadden2000mixed} adds flexibility to the original multinomial logit (MNL) model formulation \parencite{boyd1980effect} by allowing each decision-maker $n$ to  have their own preference  parameters $\beta_n$, but  constraining them to the  population density $f(\beta)$. However, this entails the assumption that the preference parameters of the decision-maker are constant throughout time and throughout different choice situations, which may be deemed too strong for certain choice modelling applications.

Individual preferences are complex and heterogeneous; they depend on different choice scenarios, and might evolve over time \parencite{castells2021novelty, hess2015intra, krueger2021evaluating}. Even when facing the same choice situation multiple times, an individual might make different decisions depending, for example, on the circumstances in the moment of choice (context). One can think of a context as an external factor that varies across individuals, as well as choice situations, and captures temporal (e.g. weather) or long-term (e.g. pandemics) circumstances. This temporal element might cause behaviour related to certain effects not to remain stable over time which can be important in explaining the variation in explanatory variables \parencite{mannering2018temporal}. This has been evident in many behavioural research fields, including transport \parencite{mannering1994temporal}, economics \parencite{meier2015temporal}, or accident research \parencite{islam2020unobserved, mannering2018temporal}.

Empirical evidence suggests that even though the major part of heterogeneity relates to inter-respondent heterogeneity, incorporating intra-respondent heterogeneity can lead to further gains in fit \parencite{hess2015intra}. Capturing this intra-heterogeneity often requires more complex model specification considerations.
\textcite{hess2009allowing} extended the MMNL model formulation, allowing for intra-respondent heterogeneity on top of the inter-respondent heterogeneity and assuming additional random variation around the mean taste across multiple choice scenarios for the same individual. This model outperforms the standard multinomial logit model in terms of prediction accuracy \parencite[in- and out-of-sample;][]{danaf2019online, xie2020personalized}, but \textcite{krueger2021evaluating} found only minor improvement when compared to the mixed multinomial logit model. \textcite{becker2018bayesian} further provided a Bayesian treatment of this intra-inter heterogeneity approach and proposed a Markov-chain Monte Carlo (MCMC) procedure for performing inference. \textcite{Danaf2020} extended this procedure by relaxing the constraint of normality assumptions. However, these approaches do not allow for systematic variations in preference parameters as a function of contextual variables.

A further approach to model heterogeneity is based on classes of individuals, with homogeneous preferences within each class. The most common example is the Latent Class Choice Model (LCCM); see, e.g. \textcite{greene2003latent} or \textcite{hess2014latent} for a discussion. A similar idea was employed in a collaborative learning framework with time-varying parameters in the context of personal recommendations \parencite{zhu2020online}. Unlike for LCCM, where an individual is perceived as belonging to one class, the individual's membership vector in a more flexible collaborative model represents a combination of multiple preference patterns identified by the model, enabling personalisation of preferences. For a detailed comparison between these two approaches, we refer to \textcite{zhu2020online}. Moreover, the collaborative learning approach overcomes the limitation of inter-intra heterogeneity model, where personalised predictions and recommendations are not possible because the individual parameters for each choice situation are not estimated.

With the developments in computational hardware, Bayesian approaches to choice modelling have been gaining research interest. \textcite{Congdon2009} elaborated on the theory and specified the task to route choice modelling. \textcite{train2001comparison} compared the Bayesian approach to mixed multinomial logit with Maximum Likelihood Simulation, an experiment repeated and extended by \textcite{Elshiew2017}. When using non-informative priors, estimates in both approaches are similar, especially for large datasets \parencite{huber2001similarity, congdon2007bayesian}. However, utilising the Bayesian approach allows researchers to include more information within the estimation procedure, thus improving the behavioural explanation of the model. It also gives a possibility to obtain full posterior distributions over the model parameters (including the individual-specific taste parameters) and take advantage of modern approaches, for example utility generation \parencite{rodrigues2020bayesian} or inference \parencite{rodrigues2022scaling}. Further extensions of the classical mixed logit model are possible using Variational Bayes for posterior inference, as shown in \textcite{Krueger2019}, where a method for including unobserved inter- and intra-individual heterogeneity in behaviour was derived.
%{\tr Technical} More recently, the rapid development in Machine Learning has resulted in many technical publications, related to e.g. the prior for the covariance matrix in MMNL \parencite{Akinc2018} or the feasibility of Bayesian inference for MMNL for large datasets \parencite{Filipe2020}.

A substantial effort has been made in extending discrete choice models with machine learning frameworks, with recent papers suggesting a joint perspective and a complementarity of these two approaches \parencite{wang2021comparing, salas2022systematic}. An example of this synergy is leveraging neural networks for the utility specification \parencite["Neural-embedded Discrete Choice Model",][]{han2019neural} through representation learning, and thus enhancing the capability to capture the inter-respondent heterogeneity \parencite{sifringer2020enhancing, han2022neural, van2019artificial}. Despite the improved prediction accuracy compared to traditional DCM methods \parencite{salas2022systematic}, these methods do not focus on improving the predictability on the individual level \parencite{krueger2021evaluating}. Thus, previous approaches had limited practicality, and efforts should be made to make these more advanced and non-linear mechanisms more applicable and interpretable in diverse settings.

In this work, we propose the context-aware Bayesian mixed multinomial logit (C-MMNL) model which allows to model context-dependent intra-respondent heterogeneity by introducing a neural network that maps contextual information to interpretable shifts in the preference parameters of the decision-maker in a Bayesian mixed multinomial logit framework. This framework allows for a joint treatment of both discrete and continuous variables and is able to capture non-linear interactions between them. Additionally, the neural network can extrapolate the behaviour of individuals to unseen context situations by leveraging information from other individuals. By making use of Stochastic Variation Inference (SVI) and GPU-hardware acceleration, we are able to handle a large-scale revealed preference (RP) dataset for bicycle route choice modelling. Lastly, despite relying on a black-box function approximator (neural network), whose non-interpretabilitable mechanism was subject to criticism in some studies, \parencite[e.g. ][]{salas2022systematic}, we show that the proposed model is highly interpretable and preserves the links to economic theories of the original MMNL. The C-MMNL model improves the individual (conditional) and general (unconditional) predictions on a hold-out sample over variations of the traditional MMNL with interaction terms. By introducing context shifts that are consistent across individuals, the estimates are not as dependent on the sample structure (number of observations per individual) as in the case of inter-intra models, where the variation of preferences for a given individual highly depends on the panel structure.

The remainder of this paper consists of four sections. Section \ref{sec:Method} briefly describes the standard MMNL model from the Bayesian perspective and introduces the proposed C-MMNL model as an extension to the MMNL model. Section \ref{sec:caseStudies} provides an extensive simulation study illustrating the estimation of the C-MMNL model and the interpretation of the results. Section \ref{sec:bicycleRCM} estimates a large-scale bicycle route choice model with the proposed method. Section \ref{sec:conclusion} concludes the paper.

\section{Method}
\label{sec:Method}
%\footnotesize{\tr A Theory section should extend, not repeat, the background to the article already dealt with in the Introduction and lay the foundation for further work. In contrast, a Calculation section represents a practical development from a theoretical basis.}

\subsection{Bayesian Mixed Multinomial Logit Model}
\label{sec:mmnl}
We consider a standard mixed multinomial logit model (MMNL) setup where on each choice occasion $t \in \{1,\dots,T\}$ a decision-maker $n \in \{1,\dots,N\}$ derives a random utility $U_{ntj} = V(\textbf{x}_{ntj},\bs\eta_n) + \epsilon_{ntj}$ from each alternative $j$ in the choice set $\mathcal{C}_{nt}$. The systematic utility term $V(\textbf{x}_{ntj},\bs\eta_n)$ is assumed to be a function of covariates $\textbf{x}_{ntj}$ and a collection of taste parameters $\bs\eta_n$, while $\epsilon_{ntj}$ is a random error term, following a type-I Extreme Value distribution. %Assuming that $\epsilon_{ntj}$ follows a type-I Extreme Value distribution, $\epsilon_{ntj} \sim \mbox{EV}(0,1)$, leads to the standard Multinomial Logit Model (MNL) kernel \textparencite{mcfadden1973conditional}, according to which the probability of the decision-maker $n$ selecting alternative $j$ is given by
%\begin{align}
%p(y_{nt} = j|\textbf{x}_{ntj},\bs\eta_n) = \frac{ e^{V(\textbf{x}_{ntj},\bs\eta_n)} }{ \sum_{k \in \mathcal{C}_{nt}} e^{V(\textbf{x}_{ntk},\bs\eta_n)} }.
%\label{eq:mnl_kernel}
%\end{align}
%For the sake of simplicity, we assume the systematic utility function, $V(\textbf{x}_{ntj},\bs\eta_n)$, to be linear-in-parameters. 
We consider the general setting under which the tastes $\bs\eta_n$ can be decomposed into a vector of fixed taste parameters $\bs\alpha \in \mathbb{R}^L$ shared across decision-makers, and individual-specific random taste parameters $\bs\beta_n \in \mathbb{R}^K$. %The systematic utility function is then defined as
%\begin{align}
%V(\textbf{x}_{ntj},\bs\eta_n) = \bs\alpha^\transpose\textbf{x}_{ntj,F} + \bs\beta_n^\transpose\textbf{x}_{ntj,R}, 
%\end{align}
%where the decomposition $\textbf{x}_{ntj}^\transpose = (\textbf{x}_{ntj,F}^\transpose, \textbf{x}_{ntj,R}^\transpose)$ is used to distinguish between covariates that pertain to the fixed parameters $\bs\alpha$ and random parameters $\bs\beta_n$, respectively. All vectors are assumed to be column vectors. 

%Following the standard MMNL model formulation
In the Bayesian framework, we assume the individual-specific taste parameters $\bs\beta_n$ to follow a multivariate normal distribution, \mbox{i.e.} $\bs\beta_n \sim \N(\bs\zeta,\bs\Omega)$. We further assume the fixed taste parameters $\bs\alpha$ and the mean vector $\bs\zeta$ to follow a multivariate normal distribution, respectively: $\bs\alpha \sim \N(\bs\lambda_0, \bs\Xi_0)$ and $\bs\zeta \sim \N(\bs\mu_0, \bs\Sigma_0)$. As for the covariance matrix $\bs\Omega$, we decompose our prior into a scale and a correlation matrix as follows: $\bs\Omega = \mbox{diag}(\bs\tau) \times \bs\Psi \times  \mbox{diag}(\bs\tau)$, where $\bs\Psi$ is a correlation matrix and $\bs\tau$ is the vector of coefficient scales \parencite{gelman2006data, barnard2000modeling}. For the components of the scale vector $\bs\tau$ we employ a vague half-Cauchy prior, \mbox{e.g.} $\tau_k \sim \mbox{half-Cauchy}(10)$, while for the correlation matrix - a LKJ prior \parencite{lewandowski2009generating}, such that $\bs\Psi \sim \mbox{LKJ}(\nu)$. The hyper-parameter $\nu$ directly controls the amount of correlation favoured by the prior.

The generative process of the MMNL model can be summarised as follows:
\begin{enumerate}
    \item Draw fixed taste parameters $\boldsymbol\alpha \sim \mathcal{N}(\boldsymbol\lambda_0, \boldsymbol\Xi_0)$ 
    \item Draw mean vector $\boldsymbol\zeta \sim \mathcal{N}(\boldsymbol\mu_0, \boldsymbol\Sigma_0)$
    \item Draw scales vector $\boldsymbol\theta \sim \mbox{half-Cauchy}(\boldsymbol\sigma_0)$
    \item Draw correlation matrix $\boldsymbol\Psi \sim \mbox{LKJ}(\nu)$
    \item For each decision-maker $n \in \{1,\dots,N\}$
    \begin{enumerate}
        \item Draw random taste parameters $\boldsymbol\beta_n \sim \mathcal{N}(\boldsymbol\zeta,\boldsymbol\Omega)$
        \item For each choice occasion $t \in \{1,\dots,T\}$
        \begin{enumerate}
            \item Draw observed choice $y_{nt} \sim \mbox{MNL}(\boldsymbol\eta_n, \textbf{X}_{nt})$
        \end{enumerate}
    \end{enumerate}
\end{enumerate}
where $\boldsymbol\Omega = \mbox{diag}(\boldsymbol\theta) \times \boldsymbol\Psi \times  \mbox{diag}(\boldsymbol\theta)$ and $\bs\eta_n = [\boldsymbol\alpha, \boldsymbol\beta_n]$. \\

\subsection{Context-aware Bayesian Mixed Multinomial Logit Model}
\label{sec:CMMNL}
%The MMNL allows for modelling heterogeneity among decision makers, keeping them constant for all choice occasions. It would be, however, reasonable to assume that individual preference parameters are dependent on the context of the choice situation. Therefore, the question arises on how to extend the MMNL model to allow non-linear dependencies on context variables without compromising on linearity of the utility function form. 
We present the idea of the context-aware Bayesian mixed multinomial logit (C-MMNL) model, where the context information is included in the form of an easily interpretable context-specific bias term $\bs\mu_t$, a non-linear function of the context information that shifts the preference parameters of each individual $n$ in each choice occasion $t$, i.e. $\bs\eta_{nt} = \bs\eta_{n} + \bs\mu_t$, where $\bs\eta_{n} = [\bs\alpha, \bs\beta_n]$. The adjustment term $\bs\mu_t$ is assumed to be determined by a neural network that takes as input the context information $\textbf{c}_t$, i.e.: $\bs\mu_{t} = \mbox{NNet}_{\bs\theta_{\mbox{\tiny NN}}}(\textbf{c}_t)$. In order to share statistical strength across individuals, we assume that all individuals shift their preference parameters in the same way when faced with a given choice context $\textbf{c}_t$, and therefore the parameters of the neural network, $\bs\theta_{\mbox{\tiny NN}}$, are shared for all individuals. However, we note that, if this assumption is considered too strong for some applications, one can relax it by allowing the neural network to also take into account, for example, the socio-demographic characteristics of the decision-maker. This would allow for complex interactions between the latter and the context information $\textbf{c}_t$. Figure~\ref{fig:pgm} shows the graphical model representation of the proposed C-MMNL model.

The generative process of the proposed C-MMNL model can then be summarised as follows, where the main changes to the generative process assumed by the original MMNL model (described in Section~\ref{sec:mmnl}) have been highlighted:
\begin{figure}[t!]
\centering
\begin{tikzpicture}[x=1.7cm,y=1.8cm]
  % Nodes
  \node[obs]    			(y)      {$y_{n,t}$} ; %
  \node[obs, above=of y]  	(x)      {$\bx_{n,t}$} ; %
  \node[latent, left=of y, draw=coolred]  	(betac)      {\color{coolred}$\bs\eta_{n,t}$} ; %
  \node[latent, left=of betac]  	(beta)      {$\bs\beta_n$} ; %
  \node[latent, left=of beta, yshift=1.2cm]  	(mu)      {$\bs\zeta$} ; %
    \node[latent, left=of beta, yshift=0.0cm]  	(psi)      {$\bs\Psi$} ; %
  \node[latent, left=of beta, yshift=-1.2cm]  	(sigma)      {$\bs\Omega$} ; %
  \node[latent, above=of mu, yshift=-0.8cm]  	(alpha)      {$\bs\alpha$} ; %
  \node[obs, above=of y, yshift=1.2cm, draw=coolred]  	(c)      {\color{coolred}$\textbf{c}_t$} ; %
  \node[latent, left=of x, yshift=1.2cm, draw=coolred]  	(muc)      {\color{coolred}$\bs\mu_{t}$} ; %
  \node[latent, left=of muc, draw=coolred]  	(theta)      {\color{coolred}$\bs\theta_{\text{NN}}$} ; %
  \edge {alpha} {betac} ; %
  \edge {x} {y} ; %
  \edge {beta} {betac} ; %
  \edge {psi} {beta} ; %
  \edge[draw=coolred] {betac} {y} ; %
  \edge {mu} {beta} ; %
  \edge {sigma} {beta} ; %
  \edge[draw=coolred]  {muc} {betac} ; %
  \edge[draw=coolred]  {c} {muc} ; %
  \edge[draw=coolred]  {theta} {muc} ; %
    \plate {inplate} { %
    (c)
    (betac)
    (x)
    (y)%
  } {$T$}; %
    \plate {explate} { %
    (inplate.south east)%(inplate.north west)
    (beta)
    (x)
    (y)%
  } {$N$}; %
\end{tikzpicture}
  \caption{Graphical model representation of the proposed C-MMNL model, where the key changes to the original MMNL model are highlighted in blue.}
  \label{fig:pgm}
\end{figure}
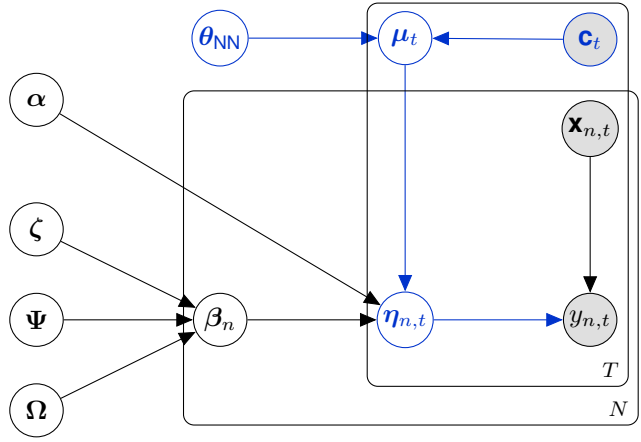

\begin{enumerate}
    \item Draw fixed taste parameters $\boldsymbol\alpha \sim \mathcal{N}(\boldsymbol\lambda_0, \boldsymbol\Xi_0)$
    \item Draw mean vector $\boldsymbol\zeta \sim \mathcal{N}(\boldsymbol\mu_0, \boldsymbol\Sigma_0)$
    \item Draw scales vector $\boldsymbol\theta \sim \mbox{half-Cauchy}(\boldsymbol\sigma_0)$
    \item Draw correlation matrix $\boldsymbol\Psi \sim \mbox{LKJ}(\nu)$
    \item For each choice occasion $t \in \{1,\dots,T\}$
        \begin{enumerate}
            \item \textbf{Determine context-specific shift term $\bs\mu_{t} = \text{NNet}_{\bs\theta_{\mbox{\tiny NN}}}(\textbf{c}_t)$}
        \end{enumerate}
    \item For each decision-maker $n \in \{1,\dots,N\}$
    \begin{enumerate}
        \item Draw random taste parameters $\boldsymbol\beta_n \sim \mathcal{N}(\boldsymbol\zeta,\boldsymbol\Omega)$
        \item For each choice occasion $t \in \{1,\dots,T\}$
        \begin{enumerate}
            \item \textbf{Compute context-adjusted taste parameters: $\bs\eta_{nt}=\bs\eta_n + \bs\mu_{t}$, with $\bs\eta_n = [\boldsymbol\alpha, \boldsymbol\beta_n]$}
            \item Draw observed choice $y_{nt} \sim \mbox{MNL}(\boldsymbol\eta_{nt}, \textbf{X}_{nt})$
        \end{enumerate}
    \end{enumerate}
\end{enumerate}
where $\boldsymbol\Omega = \mbox{diag}(\boldsymbol\theta) \times \boldsymbol\Psi \times  \mbox{diag}(\boldsymbol\theta)$. Kindly notice that we opt not to place a prior distribution on the neural network parameters $\bs\theta_{\mbox{\tiny NN}}$. Instead, the latter are treated as point parameters in the model to be estimated using a type-II maximum likelihood approach (also commonly referred to as ``empirical Bayes"). 

Letting $\bz = \{\bs\alpha, \bs\zeta, \bs\theta, \bs\Psi, \bs\mu_{1:T}, \bs\beta_{1:N}\}$ denote the set of all latent variables in the C-MMNL model, the joint distribution $p(\bz,\by_{1:N}|\bs\theta_{\mbox{\tiny NN}})$ factorises as:
\begin{align}
    p(\bz,\by_{1:N}|\bs\theta_{NN}) &= 
    p(\bs\alpha|\bs\lambda_0,\bs\Xi_0) \, p\left(\bs\zeta|\bs\mu_0, \bs\Sigma_0\right) \, p\left(\bs\theta|\bs\sigma_0\right) \,
    p\left(\bs\Psi|\nu\right)
    \left(\prod_{t=1}^{T} p\left(\bs\mu_t|\textbf{c}_t,\bs\theta_{\mbox{\tiny NN}}\right) \right) \nonumber\\
    &\times
    \prod_{n=1}^{N}
    p(\bs\beta_n|\bs\zeta,\bs\theta,\bs\Psi)
    \prod_{t=1}^{T}
    p\left(y_{nt}|\bX_{nt},\bs\eta_{nt}\right),
\end{align}
where, in our particular formulation of the model, we assume that $p\left(\bs\mu_t|\textbf{c}_t,\bs\theta_{\mbox{\tiny NN}}\right) \triangleq \delta(\bs\mu_t - \text{NNet}_{\bs\theta_{\mbox{\tiny NN}}}(\textbf{c}_t))$, with $\delta$ being a Dirac delta function. 

Our goal is then to use Bayesian inference to compute the posterior distribution of $\bz$ given a dataset of observed choices: $p(\bz|\by_{1:N},\bs\theta_{\mbox{\tiny NN}})$, while simultaneously finding a maximum likelihood estimate for the neural networks parameters $\bs\theta_{\mbox{\tiny NN}}$. Unfortunately, exact Bayesian inference in this model is intractable and therefore we must resort to approximate inference methods. Concretely, in Section~\ref{sec:estimation} we describe a Stochastic Variational Inference (SVI) algorithm for performing Bayesian inference in the proposed C-MMNL model. 

%\subsubsection{Context MMNL as a generalisation of MMNL}
%{\tr Filipe's derivation}

\subsection{Neural network}
The neural network with parameters $\bs\theta_{\mbox{\tiny NN}}$ introduced in the previous section takes as input a vector $\textbf{c}_t$, describing the context corresponding to the choice occasion $t$, and outputs a vector $\bs\mu_t \in \mathbb{R}^{L+K}$, corresponding to the context-dependent adjustments/shifts to the fixed taste parameters $\bs\alpha \in \mathbb{R}^L$ and random taste parameters $\bs\beta_n \in \mathbb{R}^K$. We propose to approximate this mapping using a neural network architecture consisting of fully-connected layers with the hyperbolic tangent activation function (tanh) for the hidden layers and a linear activation for the output layer. The number of the required hidden layers naturally depends of the complexity of the mapping. In practice, we found a single hidden layer to be sufficient for the datasets considered in the case studies in sections~\ref{sec:caseStudies} and \ref{sec:bicycleRCM}.

Given the flexibility of the neural network to capture complex patterns in the context data and their non-linear relation with the context-dependent shifts/adjustments of the taste parameters of a decision-maker $n$, the proposed C-MMNL approach offers several key advantages over traditional approaches to include context information in the utility functions. First, it naturally supports for both continuous and discrete variables, as well as complex non-linear interactions between both types of variables. As a consequence, non-linear shifts in taste parameters that result from context variables (such as amount of rain, see Section \ref{sec:bicycleRCM}) can be captured. Secondly, each specification of the context is considered jointly as a whole by the neural network, rather than each variable being considered independently. The neural network then leverages observed choice data from multiple contexts to learn to extrapolate across different combinations of the context variables and their mapping to shifts in taste parameters. Therefore, even if one has never seen a particular combination of the context variables in the data, the neural network is expected to learn to non-linearly interpolate across contexts to estimate its effect on the taste parameters. Lastly, since the parameters $\bs\theta_{\mbox{\tiny NN}}$ of the neural network are shared across all decision-makers, it can leverage information from other decision-makers to make inferences about the effect of a particular context on the taste parameters of a given decision maker. In doing so, we can leverage the ability of neural networks to generalise across observations and use it to infer the effect of a particular context, even if we have never observed that decision-maker make choices in that context. Additionally, one can further supplement the context description vector $\textbf{c}_t$ with socio-demographic information about the decision-maker in order to make the generalization across decision-makers richer and more personalised. 

\subsection{Estimation}
\label{sec:estimation}
Given that exact Bayesian inference in the proposed C-MMNL model is intractable, we propose a Stochastic Variational Inference (SVI) procedure similar to one described in \textcite{rodrigues2020bayesian}. Letting $\bz$ denote the set of all latent variables in the C-MMNL model, and $\bs\theta_{\mbox{\tiny NN}}$ denote the parameters of neural network, our goal is to compute the posterior distribution of $\bz$ and find point estimates for the neural network parameters $\bs\theta_{\mbox{\tiny NN}}$ given a dataset of observed choices. In Variational Inference (VI), we consider a tractable family of distributions $q(\bz|\bs\phi)$ parameterised by $\bs\phi$, which are referred to as the variational parameters. The goal of VI is then to find the values of $\bs\phi$ that make the variational distribution $q(\bz|\bs\phi)$ as close as possible to the true posterior distribution $p(\bz|\by_{1:N},\bs\theta_{\mbox{\tiny NN}})$, thereby effectively reducing the inference problem into an optimization problem. Following the theory of VI \parencite{jordan1998introduction}, this can be achieved by considering a lower bound on the model evidence, which in the case of C-MMNL model is given by
\begin{align}
\label{eq:elbo}
    \log p(\by|\bs\theta_{\mbox{\tiny NN}}) \geq \mathbb{E}_q[\log p(\by,\bz|\bs\theta_{\mbox{\tiny NN}})] - \mathbb{E}_q[\log q(\bz|\bs\phi)] = \mathcal{L}(\bs\phi,\bs\theta_{\mbox{\tiny NN}}).
\end{align}
Maximising the evidence lower bound $\mathcal{L}(\bs\phi,\bs\theta_{\mbox{\tiny NN}})$, or ``ELBO'' for short, w.r.t.~$\bs\phi$ is then equivalent to minimising the KL-divergence between $q(\bz|\bs\phi)$ and $p(\by|\bz,\bs\theta_{\mbox{\tiny NN}})$. 

Unfortunately, while $\nabla_{\bs\phi} \E_q [\log q(\bz|\bs\phi)]$ can generally be computed analytically given a tractable choice of approximate distribution, (\mbox{e.g.} fully-factorised, or mean-field, approximation; \textcite{Jordan1999}), computing $\nabla_{\bs\phi} \E_q [\log p(\by,\bz|\bs\theta_{\mbox{\tiny NN}})]$ exactly is infeasible for the MMNL model described in Section~\ref{sec:mmnl}, regardless of the choice of $q(\bz|\bs\phi)$. To overcome this issue, we follow a general approach for non-conjugate models using Monte Carlo gradient estimation, as proposed in \textcite{rodrigues2020bayesian} for mixed multinomial logit models. In essence, this approach consists in reparameterising the latent variables in the C-MMNL model, $\bz$, in terms of a known base distribution and a differentiable transformation. For example, if for a given latent variable $z$ in the model we assume $q(z|\bs\phi) = \N(z|\mu,\sigma^2)$, with $\bs\phi = \{\mu,\sigma\}$, we can reparameterise it as
\begin{align}
z \sim \N(z|\mu,\sigma^2) \Leftrightarrow z = \mu + \sigma\epsilon, \quad \epsilon \sim \N(0,1).
\end{align}
We can then compute gradients of an arbitrary function of $z$, $f(z)$, such as the ELBO, \mbox{w.r.t.} $\bs\phi$ by using a Monte Carlo approximation with draws from the base distribution ($\N(0,1)$ in the example above), since
\begin{align}
\nabla_{\bs\phi} \E_{q_{\bs\phi}(z)}[f(z)] \Leftrightarrow \E_{\N(\epsilon|0,1)} [ \nabla_{\bs\phi} f(\mu + \sigma\epsilon)].
\label{eq:rep_trick2}
\end{align}

As for the neural network parameters $\bs\theta_{\mbox{\tiny NN}}$, they can be estimated using a type-II maximum likelihood approach. At convergence of VI, the bound in Eq.~\ref{eq:elbo} is tight, thus making the ELBO a good proxy to the log marginal likelihood of the model $\log p(\by|\bs\theta_{\mbox{\tiny NN}})$. We can therefore find maximum likelihood estimates of the neural parameters by maximising $\mathcal{L}(\bs\phi,\bs\theta_{\mbox{\tiny NN}})$ w.r.t.~$\bs\theta_{\mbox{\tiny NN}}$. In practice, the optimization of the ELBO w.r.t.~$\bs\phi$ and $\bs\theta_{\mbox{\tiny NN}}$ is done jointly using stochastic gradient descent, where mini-batches of data are used to obtain an approximation to the gradients. As the ELBO is not a convex function, these noisy gradient approximations not only speed-up the optimization procedure, but they often turn out to help it escape poor local optima \parencite{hoffman2013stochastic}. Moreover, the use of mini-batches makes it possible to scale VI to large datasets that don't fit in memory, as is the case of the route choice dataset considered in Section~\ref{sec:bicycleRCM}. Convergence is assumed when the ELBO does not improve for a number of consecutive iterations.

The proposed C-MMNL model and the estimation procedure described above were implemented in Python and PyTorch, thereby allowing it to make use of GPU acceleration which, as shown in \textcite{rodrigues2020bayesian} and also \textcite{arteaga2022xlogit}, can lead to significant improvements in terms of computational efficiency, especially when large datasets are considered. The source code for estimating the C-MMNL model, including a few examples, is available at: [\textit{code in preparation}]. 

%Implementation
%\begin{itemize}
%    \item small subsamples of the data (mini-batches) used to approximate the ELBO gradients; it speeds up the inference; no compromising on estimation accuracy 
%    \item PyTorch (python)
%    \item GPU acceleration
%    \item environment
%    \item validated against xlogit \parencite{arteaga2022xlogit}
%\end{itemize}

\section{Simulation study}
\label{sec:caseStudies}
In this section, we apply the proposed context-aware Bayesian mixed multinomial logit model on two simulated datasets. This aims to present the mechanism and application of the C-MMNL model and to showcase how the model results are interpreted and how different context scenarios can be analysed. Furthermore, we evaluate the models in terms of parameter recovery and goodness-of-fit, and assess their predictive abilities (conditional and unconditional) in-sample and on a hold-out-sample. We compare the results of the C-MMNL model with the results of the standard MMNL model without context and an MMNL model where the context information is included in the form of interaction terms with the attributes of interest.

\subsection{Data generation}
\label{app:dataGeneration}
We conduct a simulation experiment that allows for controlling the sample composition and the true parameters of the model. The simulation mimics a route choice modeling framework and consists of the following steps.

\paragraph{Network} We design a small toy network, consisting of three origins/destinations: A, B, and C. Each OD-pair is linked by nine routes composed of all the combinations of links on the shortest paths to the destination. Each link in the network has four attributes assigned: $a_1$, $a_2$, $a_3$, and $a_4$. For each individual in the sample, the nodes A, B, and C in the network are a destination of a certain type: $d_1$, $d_2$, or $d_3$. These are allocated randomly with an equal probability of $\frac{1}{3}$.

\paragraph{Context} For each case study, we define the set of four possible context scenarios and we assume a multinomial distribution with equal probabilities to these combinations. For each scenario $c_t$, we define a deterministic function $\mu\left(t\right)$ returning a set of additive shifts to the mean of the base multivariate normal distribution of population parameters. Specifically, in each of the following cases, parameter $\beta_4$ for attribute $a_4$ is influenced by the context-dependent shifts to showcase certain effects in the C-MMNL model.

\paragraph{Individuals} We assume that the general population follows a multivariate normal distribution, $\boldsymbol\beta \sim \mathcal{N} (\boldsymbol\beta | \boldsymbol\eta, \boldsymbol\Sigma)$, $\boldsymbol\beta = \left[ \beta_1, \beta_2, \beta_3, \beta_4 \right]$. We define the mean vector $\boldsymbol\eta$, and the covariance matrix $\boldsymbol\Sigma$ as:
\begin{align*}
    \boldsymbol\eta & = [-10, -1, -2, 2 + \mu(t)]^T, \\
    \boldsymbol\Sigma_{ij} & =
    \begin{cases}
        0.2^2, & \text{for } i,j \in \{3,4\}, \quad i=j, \\
        0, & \text{otherwise.}
    \end{cases}
\end{align*}

We assume all parameters to be uncorrelated, i.e. $\boldsymbol\Sigma = \mathrm{diag} (\sigma_1^2, ..., \sigma_k^2)$. Some parameters may be constant across individuals (for some $j, \sigma_j^2 = 0$). Based on this distribution, we draw a sample of preference parameters for each individual to mimic the panel effect. 

The above-defined distribution is a base for both case studies in this section. The parameters referring to the attributes, for which a constant preference parameter is assumed ($\beta_1$ and $\beta_2$), are defined as fixed parameters in the models, while parameters $\beta_3$ and $\beta_4$ are treated as mixed (random) parameters in the model.

\paragraph{Observations} To draw the route choice observations for each individual, we first draw a context scenario based on the respective distribution and derive the associated shifts $\mu_t$ on the parameters $\boldsymbol\beta$ for the respective individual. Furthermore, we draw an OD pair $p$ from the network, using the following probability distribution\footnote{Probability distribution of bicycle trip purposes given by the Danish National Travel Survey: home-work, home-leisure, leisure-work \parencite{christiansen_danish_2022}.}: $P(d_1 \rightarrow d_2) = 0.46$, $P(d_1 \rightarrow d_3) = 0.24$, $P(d_2 \rightarrow d_3)= 0.3$. For each alternative $k$ of the drawn OD-pair, we construct a (linear-in-attributes) utility function with the respective distribution of parameters and the set of route attributes. Then, we calculate the logit probability for each alternative $k$ in OD pair $p$ and stack these probabilities by calculating their cumulative sum (i.e. calculate $p_{i,k} = \sum\limits_{l \leq k} P_{ilp}$). Finally, we draw a value $x$, uniformly distributed between 0 and 1, $x \sim \mathcal{U}(0,1)$. The value $k$ so that $x \in [p_{i,k-1}, p_{i,k}]$ is the chosen alternative.

\subsection{Evaluation criteria}
\label{sec:Evaluation}
We benchmark the proposed C-MMNL model against two alternative models - MMNL model and MMNL model with interaction terms. To measure how well each of the models can retrieve the assumed preference population parameters (and the shifts for the context scenarios), we employ the Root Mean Squared Error (RMSE), defined as $\text{RMSE}\left( \theta \right) = \sqrt{\frac{1}{K}\sum_{k=1}^{K} \left(\theta_k -\hat{\theta}_k\right)^2}$. Please note that we extend the definition of RMSE (which is traditionally defined to capture the difference between parameters and their estimates) to measure the difference in both the estimated mean vector $\bs\eta$ and, in the case of the C-MMNL model, the set of shifts $\bs\mu_t$. Furthermore, if the parameters (or shifts) $\theta$ are defined on a continuous domain (as in Section \ref{sec:caseStudyCont}), we approximate the difference $\theta_k - \hat{\theta}_k$ on the whole domain with a finite sum.

We aim to evaluate both the conditional (individual-level) and unconditional prediction ability of the proposed C-MMNL and the benchmark models, both in-sample and on a hold-out sample. We split each of the generated datasets into a train and a test dataset as follows:
 \begin{itemize}
      \item for the general prediction, we randomly choose 80\% of the individuals and use all their trips to form a train set; test set consists of the remaining 20\% of individuals, 
     \item for the individual prediction, we randomly choose 80\% of the trips for all individuals for a train set; the remaining 20\% of the trips for all individuals constitute a test set.
 \end{itemize}

We evaluate the fit and predictive abilities of the model utilising several criteria. First, we report the average model probability of the chosen alternative $\bar{P}_{\text{ch}}$ (as in \cite{danaf2019online, xie2020personalized, krueger2021evaluating}). We further report the Brier score loss (\cite{brier1950verification}, as in \cite{krueger2021evaluating}), which measures the mean squared difference between the predicted probability and the actual outcome, taking into account also the probabilities of the non-chosen alternatives, and is defined as:
\begin{equation}
    \text{Brier} = \frac{1}{NTJ} \sum_{n=1}^N \sum_{t=1}^T \sum_{j=1}^J \left( \mathbbm{1}_{\{ y_{nt} = j \}} - \hat{P}_{ntj}\right)^2,    
\end{equation}
where $y_{nt}$ is the observed choice in the choice situation $t$ for an individual $n$, $\hat{P}_{ntj}$ is the predicted logit probability, assigned to the alternative $j$ in that choice situation, and $\mathbbm{1}_x$ denotes the indicator function. The lower the Brier score value, the better the model performs in terms of prediction. Finally, for completeness, we also report the log-likelihood.

In each of the case studies, we perform the analysis for $N=500$ individuals and $T=10$ and $T=50$ observations per individual, respectively. All presented estimation results and values of the evaluation criteria are based on an averaged model across 100 populations. Please note that datasets for both case studies are defined and generated separately and are not to be compared across cases.

\subsection{Case study 1}
In the first case, we define two binary context variables $c_1$ and $c_2$, which results in four possible context scenarios. We define the function $\mu(c_1,c_2)$, returning shifts to the mean of the parameter $\beta_4$, as:
\begin{align*}
    \mu\left(c_1, c_2\right) =
    \begin{cases}
        0, & \text{for } c_1 = 0 \wedge c_2 = 0, \\
        1, & \text{for } c_1 = 0 \wedge c_2 = 1, \\
        2, & \text{for } c_1 = 1 \wedge c_2 = 0, \\
        3, & \text{for } c_1 = 1 \wedge c_2 = 1.\\
    \end{cases}
\end{align*}
We specify the base utility function for the choice situation $i$ of the individual $n$ as:
\begin{equation}
\label{eq:utilityBase}
    V_{n, i} = \beta_1 X_{1,i} + \beta_2 X_{2,i} + \beta_{3,n} X_{3,i} + \beta_{4,n} X_{4,i}.
\end{equation}
The MMNL model with the above utility specification serves as a benchmark model. In this simple form, the MMNL captures only the inter-respondent heterogeneity and does not separate the effect of the context on the observation level.

Incorporating this additional level of heterogeneity in the preference parameters can be achieved by extending the utility specification from Equation \ref{eq:utilityBase} by the interaction terms between the attribute $a_4$ and the context attributes $c_1$ and $c_2$:
\begin{align}
\label{eq:utilityExt}
    \tilde{V}_{n, i} = \beta_1 X_{1,i} + \beta_2 X_{2,i} + \beta_{3} X_{3,i} + \beta_{4}^{(0,0)} X_{4,i} + \beta_{4}^{(0,1)} X_{4,i} \cdot c_2 + \beta_{4}^{(1,0)} X_{4,i} \cdot c_1 + \beta_{4}^{(1,1)} X_{4,i} \cdot c_1 \cdot c_2
\end{align}
The notation for individual parameters $\beta_{3,n}$ and $\beta_{4,n}$ is simplified for readibility reasons.

Including the interaction between one network attribute $a_4$ and two binary context variables $c_1$ and $c_2$ results in an extension to the utility function by three further terms. If we had more context variables at hand and/or were interested in further interaction effects, the size of the utility function would grow immensely, making the interplay of the parameters difficult to interpret. For example, interpreting the effect of the scenario $(c_1, c_2) = (1,1)$ requires tracking down all coefficients contributing to the final parameter (in this particular case, all coefficients $\beta_{4}^{(0,0)}, \beta_{4}^{(0,1)}, \beta_{4}^{(1,0)}$, and  $\beta_{4}^{(1,1)}$).

The C-MMNL proposes an alternative approach to capturing the effect of the context attributes. Instead of including additional parameters, C-MMNL employs a neural network to output deterministic additive shifts $\mu$ on selected fixed or random parameters. It requires only the base specification of the utility function (as in Equation \ref{eq:utilityBase}) and defining the set of utility attributes that should be "interacted" with the context - all possible context scenarios result immediately from the input data. The extension to the utility function in the C-MMNL model can be formalised as follows:
\begin{equation}
\label{sec:utilityCTX}
    \dbtilde{V}_{n, i} = \beta_1 X_{1,i} + \beta_2 X_{2,i} + \beta_{3} X_{3,i} + \left( \beta_{4}^{\text{base}} + \mu_i \right) X_{4,i}.\\ 
\end{equation}

We estimate all three above-mentioned models with the respective utility specifications, and we present the results in Table \ref{tab:Case1Res}.

MMNL with interaction terms captures all four context scenarios through partial interaction components. To analyse the preference parameter for the attribute $a_4$ in a fully defined context scenario, the respective coefficients have to be summed up. For example, in the scenario $(c_1, c_2) = (1,1)$ all four interaction terms have to be summed up, resulting in the value 5.650, if we consider the non-significant parameter for $\beta_{4}^{(1,1)}$ (the nominal value of the parameter for this scenario is 5).

In the case of the proposed C-MMNL model, the results for the attribute $a_4$ (which is assumed to be affected by the context variables in the model specification) have a slightly different interpretation. For this parameter, the model first outputs a non-interpretable base parameter $\beta_{4}^{\text{base}}$. Additionally, evaluations of the neural network for each context scenario of interest result in additive shift values. The results of the C-MMNL model in Table \ref{tab:Case1Res} are presented with the scenario $(c_1, c_2) = (0,0)$ as the base scenario, and the shifts should be interpreted as additional effects to that base scenario to enable a simple comparison with the nominal shift function $\mu(c_1, c_2)$. The results indicate that the C-MMNL model is able to reproduce both the preference parameters and the shifts very well, with an almost-zero RMSE in the case of $T=50$.

\begin{table}[H] \footnotesize
\centering
\hspace*{-1cm}
\begin{tabular}{lllllllllll}
\hline \noalign{\vskip 2pt}
 &              & \multicolumn{2}{l}{MMNL}           & \multicolumn{2}{c}{MMNL with interactions}       & \multicolumn{5}{l}{C-MMNL}                                                            \\ \hline \noalign{\vskip 2pt}
                &               & \textit{}       & \textit{}        & \textit{}                     & \textit{}        &  $(0,0)$          &       & $(0,1)$          & $(1,0)$          & $(1,1)$         \\ \noalign{\vskip 2pt}
                & Parameter name & coeff. & z-value & coeff.               & z-value & coeff. & z-value & \multicolumn{3}{c}{Shifts to $\beta_{4}^{(0,0)}$} \\ \hline \noalign{\vskip 2pt}
                T = 10 &  \\
                & $\beta_1$        & -7.137          & -161.814         & -11.105                       & -189.851        & \multicolumn{1}{|l}{-10.765}         & -200.663         &                &                &                \\
                & $\beta_2$     & -0.518          & -13.270          & -1.169                        & -23.392              & \multicolumn{1}{|l}{-1.078}           & -23.642          &                &                &                \\
                & $\beta_3$    & -1.261           & -32.022          & -2.216                        & -43.067          & \multicolumn{1}{|l}{-2.127}          & -52.113          &                &                &                \\
                & $\beta_4$      & 2.285           & 57.134          & ---                      & ---        &  \multicolumn{1}{|l}{ --- } & --- \\   
                & $\beta_{4}^{(0,0)}$   & --- & ---          & 2.156                         & 40.798          & \multicolumn{1}{|l}{2.121}           & 50.854           & +1.192         & +2.133         & +3.248        \\ 
                & $\beta_{4}^{(0,1)}$  & ---  & ---                & 1.332                         & 20.570           &  \multicolumn{1}{|l}{ }               &                  &                &                &                \\
                & $\beta_{4}^{(1,0)}$   & --- & ---                & 2.269                         & 35.455           &   \multicolumn{1}{|l}{ }              &                  &                &                &                \\
                & $\beta_{4}^{(1,1)}$   & --- & ---                & {\tg -0.107} & -1.302           &  \multicolumn{1}{|l}{ }               &                  &                &                &                \\
                & $\beta_3$ (sd) & {\tg 0.089}              & 1.325               & {\tg 0.098}                            & 1.322               & \multicolumn{1}{|l}{{\tg 0.095}}              & 1.331               &                &                &                \\
                & $\beta_4$ (sd) & {\tg 2.266}              & 1.339               & {\tg 2.582}                            & 1.336               & \multicolumn{1}{|l}{{\tg 2.334}}              & 1.333               &                &                &                \\  \noalign{\vskip 0.3cm}    
                & RMSE (sd)         & \multicolumn{2}{l}{---}              & \multicolumn{2}{l}{0.468 (0.065)}                        & \multicolumn{5}{l}{0.366 (0.086)}                                                             \\  \noalign{\vskip 2pt} \hline \noalign{\vskip 2pt}
                T = 50 &  \\
                & $\beta_1$        & -6.347          & -171.106         & -10.477                       & -216.411        & \multicolumn{1}{|l}{-10.120}         & -203.193         &                &                &                \\
                & $\beta_2$     & -0.439          & -12.734          & -1.069                        & -24.273              & \multicolumn{1}{|l}{-1.011}           & -23.598          &                &                &                \\
                & $\beta_3$      & -1.099           & -30.172          & -2.095                        & -43.146          & \multicolumn{1}{|l}{-2.013}          & -52.339          &                &                &                \\
                & $\beta_{4}$      & 2.024           & 56.658          & ---                      & ---        &  \multicolumn{1}{|l}{ --- } & --- \\   
                & $\beta_{4}^{(0,0)}$   & --- & ---          & 2.068                         & 42.912          & \multicolumn{1}{|l}{2.016}           & 54.046          & +1.047         & +2.006         & +3.039        \\ 
                & $\beta_{4}^{(0,1)}$  & ---  & ---                & 1.150                         & 21.307          &  \multicolumn{1}{|l}{ }               &                  &                &                &                \\
                & $\beta_{4}^{(1,0)}$   & --- & ---                & 2.109                        & 39.632            &   \multicolumn{1}{|l}{ }              &                  &                &                &                \\
                & $\beta_{4}^{(1,1)}$   & --- & ---                & {\tg -0.057} & -0.744           &  \multicolumn{1}{|l}{ }               &                  &                &                &                \\
                & $\beta_3$ (sd) & {\tg 0.261}              & 1.328               & {\tg 0.224}                            & 1.327               & \multicolumn{1}{|l}{{\tg 0.181}}              & 1.327               &                &                &                \\
                & $\beta_4$ (sd) & {\tg 2.061}              & 1.432               & {\tg 2.032}                            & 1.341               & \multicolumn{1}{|l}{{\tg  2.011}}              & 1.373               &                &                &                \\  \noalign{\vskip 0.3cm}    
                & RMSE (sd)         & \multicolumn{2}{l}{---}              & \multicolumn{2}{l}{0.203 (0.036)}                        & \multicolumn{5}{l}{0.070 (0.031)}                                                             \\  \noalign{\vskip 2pt} \hline \noalign{\vskip 2pt}                
\end{tabular}
\caption{Results of the proposed C-MMNL model and the benchmark models (case study 1). Coefficients in gray are insignificant on the 5\% level.}
\label{tab:Case1Res}
\end{table}

We now evaluate the performance of each of the three models, in-sample and on a hold-out sample (in two prediction scenarios), with the criteria defined in Section \ref{sec:Evaluation}. Generally, the C-MMNL model performs slightly worse in-sample than the MMNL with interactions, but always performs better on a hold-out sample (both general and individual predictions). The superiority of C-MMNL over MMNL with interactions is already visible in an example with a discrete set of shifts defined on a discrete domain for the context variables. It becomes more pronounced with a non-linear effect of a context attribute defined for a continuous domain, as shown in the subsequent case study.

\begin{table}[H] \footnotesize
\centering
\hspace*{-1cm}
\begin{tabular}{lllllllllll}
\hline \noalign{\vskip 2pt}
&                      & \multicolumn{3}{l}{MMNL}    & \multicolumn{3}{l}{MMNL with interactions} & \multicolumn{3}{l}{C-MMNL} \\ \hline \noalign{\vskip 2pt}
          &                      & $\bar{P}_{\text{ch}}$ & Brier & LL        & $\bar{P}_{\text{ch}}$      & Brier & LL             & $\bar{P}_{\text{ch}}$ & Brier & LL  \\  \noalign{\vskip 2pt} \hline \noalign{\vskip 2pt}
T = 10    &                      &         &       &           &              &            &                &         &       &           \\
          & In-sample            & 0.421   & 0.064 & -3,885.66 & 0.631        & 0.039      & -2,134.41      & 0.596   & 0.044 & -2,395.42 \\
          & Hold-out (general)    & 0.398   & 0.072 & -1,448.28 & 0.553        & 0.055      & -948.995       & 0.547   & 0.054 & -939.778  \\
          & Hold-out (individual) & 0.377   & 0.075 & -1,224.63 & 0.545        & 0.057      & -801.066       & 0.54    & 0.056 & -776.302  \\  \noalign{\vskip 2pt} \hline \noalign{\vskip 2pt}
T = 50    &                      &         &       &           &              &            &                &         &       &           \\
          & In-sample            & 0.376   & 0.069 & -21,486.2 & 0.571        & 0.047      & -12,931.7      & 0.555   & 0.049 & -13,481.8 \\
          & Hold-out (general)      & 0.371   & 0.071 & -7,049.3  & 0.543        & 0.053      & -4,616.68       & 0.536   & 0.053 & -4,595.83 \\
          & Hold-out (individual) & 0.367   & 0.071 & -5,630.19 & 0.545        & 0.053      & -3,673.4        & 0.542   & 0.052 & -3,575.82 \\  \noalign{\vskip 2pt} \hline \noalign{\vskip 2pt}    
\end{tabular}
\caption{Evaluation of the proposed C-MMNL model and the benchmark models, both in-sample and on hold-out samples (case study 1).}
\label{tab:Case1Eval}
\end{table}

\subsection{Case study 2}
\label{sec:caseStudyCont}
In the second example, we introduce two nuances: first, one of the context variables $c_2$ is defined on a continuous domain $\left[0,3\right]$, and second, the function $\mu\left(c_1, c_2\right)$ has a continuous range for one context scenario. This example aims to reflect the following situation: in the exclusive presence of the context attribute $c_2$, the preference parameter for attribute $a_4$ is shifted rapidly for low positive values of context $c_2$ and further stabilises within the domain. However, when the other context attribute is present $(c_1 = 1)$, it determines the preference parameter $\beta_4$ and neutralises the effect of $c_2$. These effects can be formalised as follows:
\begin{align*}
    \mu\left(c_1, c_2\right) =
    \begin{cases}
        0, & \text{for } c_1 = 0 \wedge c_2 = 0, \\
        -2\exp{\left(-2c_2\right)} + 2, & \text{for } c_1 = 0 \wedge c_2 \in \mathopen(0,3\mathclose],  \\
        2, & \text{for } c_1 = 1 \wedge c_2 = 0, \\
        2, & \text{for } c_1 = 1 \wedge c_2 \in \mathopen(0,3\mathclose].  \\
    \end{cases}
\end{align*}

\begin{figure}[H]
    \centering
    \begin{subfigure}[b]{0.49\textwidth}
        \includegraphics[width=\textwidth]{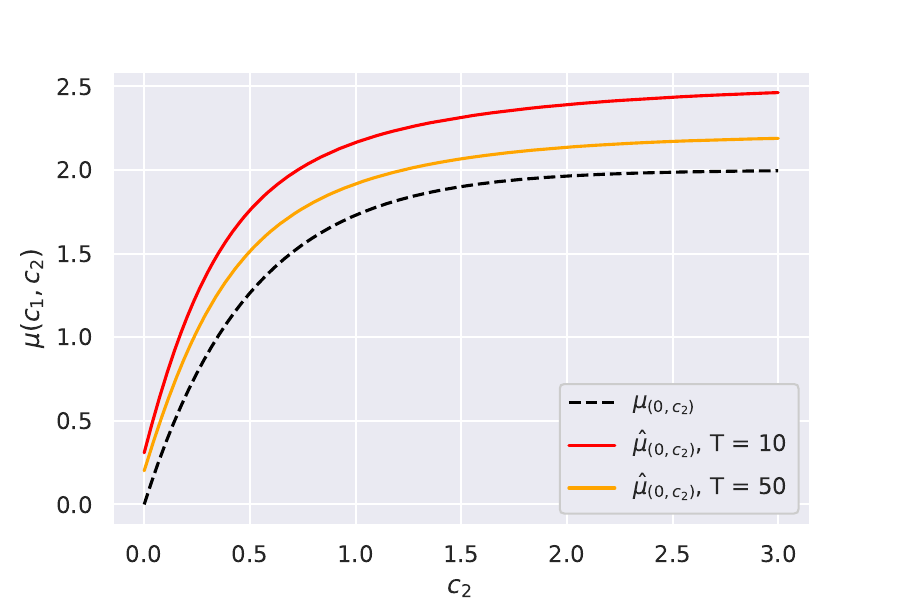}
        \caption{Shift on the parameter $\beta_{4}^{(0,0)}$ for $c_1 = 0$.}
        \label{fig:mu01Vsmu11}
    \end{subfigure}
    \hfill
    \begin{subfigure}[b]{0.49\textwidth}
        \includegraphics[width=\textwidth]{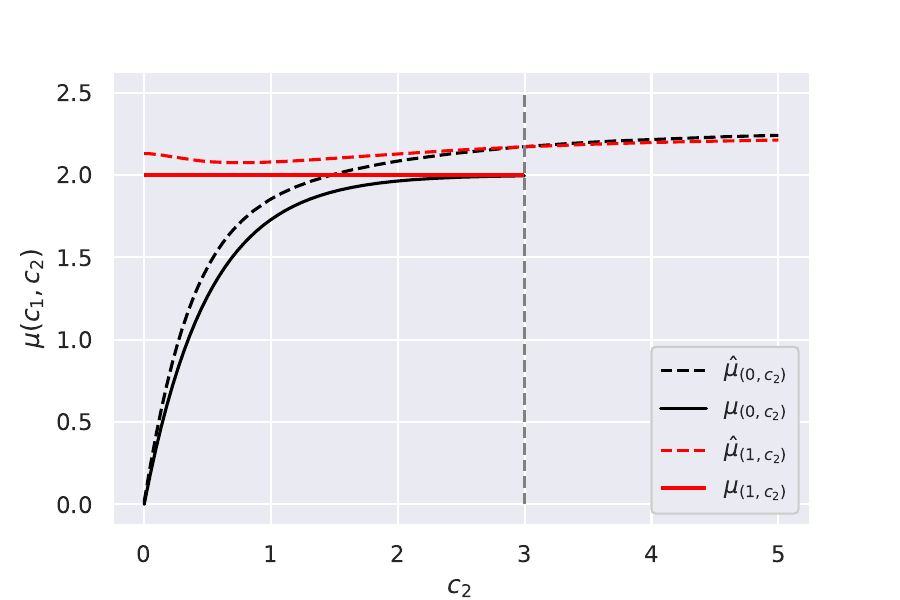}
        \caption{Shifts on the parameter $\beta_{4}^{(0,0)}$, $T=10$.}
        \label{fig:mu01_T10vsT50}
    \end{subfigure}
    \caption{Recovery of the nominal shifts on the preference parameter $\beta_{4}^{(0,0)}$ (case study 2).}
\end{figure}

Figures \ref{fig:mu01Vsmu11} and \ref{fig:mu01_T10vsT50} indicate how the neural network in the proposed C-MMNL model reproduces the nominal shape of the shifts. The development of the shift can be separated for different scenarios and the model can capture the non-linear relationship between the context attributes. Therefore, it is possible to determine to which context attribute the change in preference parameter can be attributed and to which extent. Furthermore, the neural network is able to extrapolate beyond the support of $c_2$, therefore determining the values of the shifts (and of the preference parameters) for context situations unknown to the model at the estimation stage (Figure  \ref{fig:mu01_T10vsT50}).

As a benchmark, we estimate an MMNL model with interaction terms, where we assume a simplified binary domain for the context attribute $c_2$ ($c_2=0$ is interpreted as absence, and $c_2>0$ as presence of the attribute). This model estimates discrete shifts to the preference parameter $\beta_4$, and specifically for the scenario $(c_1, c_2) = (0,1)$ it estimates a constant shift of 1.771 (1.579 for $T=50$) on the base parameter $\beta_{4}^{(0,0)}$ and does not account for the continuous development of the shift. Also, in the scenario $(c_1, c_2) = (1,1)$, the model is not able to separate the effect of each of the context attributes on the final preference parameter value.

In terms of RMSE, the proposed C-MMNL model outperforms the MMNL with interactions by far (0.882 vs. 0.386 for $T=10$ and 0.625 vs. 0.129 for $T=50$). Table \ref{tab:Case2Eval} includes the evaluation criteria for all three estimated models. Again, the C-MMNL model indicates a better prediction performance on the hold-out sample, and the magnitude of the improvement is higher than in the first case study. These improvements in the parameter recovery and the prediction performance on the hold-out sample can be mostly attributed to the difference in estimates for the context scenario on the continuous domain that C-MMNL is able to capture, as well as to the shift in case of the presence of both attributes.

\begin{table}[H] \footnotesize
\centering
\hspace*{-1cm}
\begin{tabular}{lllllllllll}
\hline \noalign{\vskip 2pt}
&                      & \multicolumn{3}{l}{MMNL}    & \multicolumn{3}{l}{MMNL with interactions} & \multicolumn{3}{l}{C-MMNL} \\ \hline \noalign{\vskip 2pt}
          &                      & $\bar{P}_{\text{ch}}$ & Brier & LL        & $\bar{P}_{\text{ch}}$      & Brier & LL             & $\bar{P}_{\text{ch}}$ & Brier & LL  \\  \noalign{\vskip 2pt} \hline \noalign{\vskip 2pt}
T = 10 &                      &         &       &           &              &            &                &         &       &           \\
       & In-sample            & 0.466   & 0.060 & -3,558.01 & 0.632        & 0.039      & -2,130.36      & 0.606   & 0.043 & -2,306.96 \\
       & Hold-out (general)    & 0.449   & 0.067 & -1,340.62 & 0.557        & 0.055      & -962.112       & 0.553   & 0.054 & -916.419  \\
       & Hold-out (individual) & 0.423   & 0.069 & -1,134.11 & 0.549        & 0.057      & -800.357       & 0.550   & 0.055 & -753.444  \\ \noalign{\vskip 2pt} \hline \noalign{\vskip 2pt}
T = 50 &                      &         &       &           &              &            &                &         &       &           \\
       & In-sample            & 0.422   & 0.064 & -19,816.6 & 0.571        & 0.048      & -13,047.2      & 0.562   & 0.049 & -13,112.7 \\
       & Hold-out general      & 0.419   & 0.066 & -6,557.06 & 0.544        & 0.054      & -4,684.66      & 0.540   & 0.053 & -4,485.23 \\
       & Hold-out (individual) & 0.414   & 0.066 & -5,200.82 & 0.546        & 0.053      & -3,716.15      & 0.549   & 0.051 & -3,478.83 \\  \noalign{\vskip 2pt} \hline \noalign{\vskip 2pt}    
\end{tabular}
\caption{Evaluation of the proposed C-MMNL model and the benchmark models, both in-sample and on hold-out samples (case study 2).}
\label{tab:Case2Eval}
\end{table}

\section{Large-scale case study}
\label{sec:bicycleRCM}
This section presents an application of the proposed C-MMNL model to a real-world case from the transport domain - a large-scale bicycle route choice model. We estimate the C-MMNL model on a dataset of 119,448 trips from 8,555 individuals with 14 parameters and three contextual attributes. The model is evaluated with criteria from Section \ref{sec:caseStudies} and benchmarked against MMNL model without context attributes. This example illustrates the flexibility of the proposed approach and its scability to large datasets.

\subsection{Data}
\label{sec:data}
In this example, we utilise a large-scale, crowdsourced dataset of GPS trajectories collected from advanced bicycle head protection devices from the company Hövding\footnote{\url{https://www.hovding.com}}. The initial dataset covers the entire Copenhagen metropolitan area in the period from the 16\textsuperscript{th} September 2019 until 31\textsuperscript{st} May 2021 and consists
of 365,813 trips from 10,049 individuals. We apply a combination of steps to process the GPS trajectories based on \textcite{lissner2021facing} and \textcite{Schuessler2009}. The data is map-matched to a highly disaggregated bicycle network based on Open Street Map (OSM\footnote{\url{https://www.openstreetmap.org}}) and the choice set is generated by the ``categorical'' approach from \textcite{Rasmussen2021CRBAM}. To create a set of possible determinants influencing cyclists' route choice behaviour, we process and manipulate the attributes of a highly detailed bicycle network of the Copenhagen metropolitan area.  The analysis is based on a highly detailed bicycle network (420,973 directed links and 324,492 nodes) and the model takes into account the following characteristics: length, elevation gradient, bicycle infrastructure, surface type, landuse, and the allowed direction of traffic.

For a detailed description of all data sources and pre-processing algorithms utilised in this study, we refer to \textcite{lukawska2023routechoice}. The final dataset for route choice modelling is restricted to a maximum of 50 trips per cyclist to avoid sample imbalance and consists of 119,448 trips made by 8,555 users.

\textcite{ton2017people} listed contextual trip attributes potentially relevant for the case of route choice behaviour of cyclists, such as weather, daylight, trip purpose, or cycling season. In some of the existing bicycle route choice studies, the models accounted for additional trip-related context variables, for example, weather \parencite{Prato2018}, trip time \parencite{khatri2016modeling, ton2017people, dane2019route, shah2021different, Prato2018} or trip purpose \parencite{hood2011gps, Broach2012a, bernardi2018modelling, dane2019route}. The studies found, for example, that cyclists on commute trips and on trips during peak hours are relatively less likely to detour from the shortest paths, compared to cyclists on other utilitarian trips or trips during off-peak hours, respectively. The vast majority of the models were estimated using the standard Maximum Likelihood Estimation procedure. The presented C-MMNL model for bicycle route choice is an addition to this line of research. We propose a rather simple model formulation that is able to handle complex behavioural patterns in the data and is easily scalable to large datasets.

In order to gain a deeper understanding of cyclists' behaviour in different contexts, we draw inspiration from existing literature and incorporate a range of contextual attributes. The context variables attributed to a trip are based on the value at the first point of the GPS trajectory and are presented in Table \ref{tab:ctxRouteChoice} together with their source.

\begin{table}[H]
    \centering
    \hspace*{-1cm}
    \begin{tabular}{lll}
    \noalign{\vskip 2pt} \hline \noalign{\vskip 2pt}
        Context variable & Source & Description\\
        \noalign{\vskip 2pt} \hline \noalign{\vskip 2pt}
        %peak hours & timestamp & 6-9am/3-6pm, Monday to Friday \\[1em]
        
        darkness & package \textit{Sun} in Python & based on sunrise/sunset time \\[1em]
        
        utilitarian trips & heuristics from \textcite{lissner2021facing} & utilitarian or leisure purpose \\[1em]
        
        amount of rain & OpenWeatherMap & \makecell[l]{amount of rain within the last\\hour before the start of a trip} \\
        \noalign{\vskip 2pt} \hline \noalign{\vskip 2pt}
    \end{tabular}
    \hspace*{-1cm}
    \caption{Context variables in the large-scale bicycle route choice model.}
    \label{tab:ctxRouteChoice}
\end{table}

\subsection{Bicycle route choice model}
\paragraph{Model specification}
We assume that each of the attributes derived from the network contributes to the utility function linearly, i.e. the systematic utility term is assumed to be a linear function of network attributes and a collection of taste parameters. The utility expression associated with alternative $j$ in choice situation $t$ for individual $n$ is defined as:
\begin{align}
    U_{ntj} = \bs\eta_n\bm{x}_{ntj} + \beta_{\text{PS}} \ln{\text{PS}_{ntj}} + \epsilon_{ntj},
\end{align}
where $\bs\eta_n = [\boldsymbol\alpha, \boldsymbol\beta_n]$ is a collection of both fixed and mixed parameters,  $\bm{x}_{ntj}$ is a set of network attributes, $\text{PS}_{ntj}$ is the path-size coefficient accounting for the overlap between alternatives \parencite{Frejinger2007}, and $\epsilon_{ntj}$ is the random error (iid, EV1-distributed).

In this example, we study the influence of the three contextual attributes (see Section \ref{sec:data}) on five network attributes: length, protected bicycle tracks, gravel surface, areas near water, and low-rise urban areas. As mentioned in the simulation study in Section \ref{sec:caseStudies}, the utility specification for the C-MMNL model is not extended by further terms and the model requires only the interactions of interest to be predefined.
    
\paragraph{Model results}
Table \ref{tab:model} presents the results and evaluation of both C-MMNL model and the benchmark MMNL model. For the C-MMNL model, it includes four scenarios, determined by all possible combinations of the two binary context attributes "darkness" and "utilitarian trips". The values of the coefficients and the shifts represent the average marginal effects w.r.t. the third context variable "amount of rain". Please note that the analysis of these shifts in isolated, fully-defined scenarios is also possible, e.g. if an analysis of the marginal effects in a heavy rain scenario was to be conducted.

\begin{table}[H] \footnotesize
    \centering
    \hspace*{-1cm}
    \begin{tabular}{llllllll} 
    \noalign{\vskip 2pt} \hline \noalign{\vskip 2pt}
    & MMNL & & \multicolumn{4}{l}{C-MMNL} \\
    \noalign{\vskip 2pt} \hline \noalign{\vskip 2pt}
        & &  & \multicolumn{2}{l}{\makecell{Dark - \\Util -}}
          & \makecell{Dark - \\Util +}
          & \makecell{Dark + \\Util -}
          & \makecell{Dark + \\Util +}\\ \noalign{\vskip 2pt}
        Parameter name & coeff. & z-value & 
        coeff. & z-value & \multicolumn{3}{c}{Shifts to the base} \\ 
    \noalign{\vskip 2pt} \hline \noalign{\vskip 2pt}
    \textbf{Trip characteristics} & & \multicolumn{1}{l|}{} & & &\\
    \tabindent Length & -8.730 & \multicolumn{1}{l|}{-26.725} & -8.182 & -24.899 & -2.058 & -0.285 & -2.272\\
    \tabindent ln(Path-Size) & 0.355 & \multicolumn{1}{l|}{3.778} & 0.433 & 4.799 &  \\
    \textbf{Elevation gradient} & & \multicolumn{1}{l|}{} & & &\\
    \tabindent Flat or downhill & ref. & \multicolumn{1}{l|}{---} & ref. & --- &\\
    \tabindent Steep uphill (10 --- 35 m/km) & -0.117 & \multicolumn{1}{l|}{-2.640} & -0.102 & -2.279 &  \\
    \tabindent Very steep uphill  ($>$ 35 m/km) & -0.168 & \multicolumn{1}{l|}{-4.164} & -0.159 & -3.901 &  \\
    \textbf{Bicycle infrastructure} & & \multicolumn{1}{l|}{} & & &\\
    \tabindent No bicycle infrastructure & ref. & \multicolumn{1}{l|}{---} & ref. & --- &\\    
    \tabindent Painted bicycle lanes     & 1.143 & \multicolumn{1}{l|}{24.285} & 1.188 & 21.018 &  \\    
    \tabindent Protected bicycle tracks  & 1.658 & \multicolumn{1}{l|}{18.777} & 1.605 &  29.857 & +0.294 & -0.012 & +0.346\\   
    \textbf{Surface type} & & \multicolumn{1}{l|}{} & & &\\
    \tabindent Asphalt & ref. & \multicolumn{1}{l|}{---} & ref. & --- &\\    
    \tabindent Cobblestones & -3.101 & \multicolumn{1}{l|}{-23.978} & -3.138 & -26.975 \\   
    \tabindent Gravel & -2.699 & \multicolumn{1}{l|}{-17.822} &  -2.532 &  -19.940 & -0.181 & -0.342 & -0.454\\   
    \textbf{Wrong way} & -1.867 & \multicolumn{1}{l|}{-16.274} & 1.988 & -13.914  \\ 
    \textbf{Land-use (right-hand side)} & & \multicolumn{1}{l|}{} & & &\\
    \tabindent High-rise urban areas & ref. & \multicolumn{1}{l|}{---} & ref. & --- &\\    
    \tabindent Green areas & {\tg 0.105} & \multicolumn{1}{l|}{1.578} & 0.206 & 3.039 \\ 
    \tabindent Areas near water & 0.756 & \multicolumn{1}{l|}{10.660} & 1.019 & 10.832 & -0.133 & -0.421 & -0.285\\
    \tabindent Industrial areas & {\tg -0.068} & \multicolumn{1}{l|}{-0.919} &  {\tg 0.014} & 0.170\\ 
    \tabindent Low-rise urban areas & -0.505 & \multicolumn{1}{l|}{-6.952} & -0.366 & -4.155 & +0.071 & -0.088 & -0.027\\ 
    \tabindent Open landscape & {\tg 0.114} & \multicolumn{1}{l|}{1.678} & 0.265 & 3.453 \\ 
    \tabindent Green areas (sd) & 1.183 & \multicolumn{1}{l|}{20.727} &  1.228 & 20.848\\ 
    \tabindent Areas near water (sd) & 2.543 & \multicolumn{1}{l|}{{\tg 1.871}} & {\tg 2.656} & 1.804 \\
    \tabindent Industrial areas (sd) & 4.006 & \multicolumn{1}{l|}{2.233} & 3.817 & 2.219 \\ 
    \tabindent Low-rise urban areas (sd) & 5.030 & \multicolumn{1}{l|}{2.509} & 5.032 & 2.579 \\ 
    \tabindent Open landscape (sd) & 5.786 & \multicolumn{1}{l|}{2.903} & 5.823 &  2.955 \\ 
    \noalign{\vskip 2pt} \hline \noalign{\vskip 2pt}
    No. of observations: & \multicolumn{2}{l|}{119,448} & \multicolumn{4}{l}{119,448}\\
    No. of individuals: & \multicolumn{2}{l|}{8,555} & \multicolumn{4}{l}{8,555}\\
    No. of utility parameters: & \multicolumn{2}{l|}{14} & \multicolumn{4}{l}{14}\\
    Log-likelihood: & \multicolumn{2}{l|}{-224,202} & \multicolumn{4}{l}{-207,102} \\
    Average prob. of the choice $\bar{P}_{\text{ch}}$: & \multicolumn{2}{l|}{0.242} & \multicolumn{4}{l}{0.264}\\
    Brier score & \multicolumn{2}{l|}{0.024} & \multicolumn{4}{l}{0.023}\\
    \% of correct predictions: & \multicolumn{2}{l|}{38.43\%} & \multicolumn{4}{l}{40.66\%}\\
    Estimation time (GPU) [s]: & \multicolumn{2}{l|}{2,329} & \multicolumn{4}{l}{2,792}\\
    \noalign{\vskip 2pt} \hline \noalign{\vskip 2pt}
    \end{tabular}
    \hspace*{-1cm}
    \caption{Estimated bicycle route choice models.  Coefficients in gray are insignificant on the 5\% level. Parameters and shifts in the C-MMNL model represent the average marginal effect w.r.t. the context variable "amount of rain".}
     \label{tab:model}
\end{table}

The results for the C-MMNL model are included in the form of additive shifts to the base scenario (where the value of both binary context variables equals 0). The analysis of the single scenarios does not require a reestimation of the model - instead, the model learns the structure from the data and the results for each of the scenarios are simply obtained by performing additional forward passes through the neural network. By applying the C-MMNL model, we avoid adding an extra parameter for each context interaction of interest. As discussed earlier, models with many contextual variables tend to be complex w.r.t. the number of degrees of freedom and the results might be challenging to interpret.

Both models reveal high heterogeneity for land-use variables, which is consistent with the previous bicycle route choice study from Copenhagen \parencite{Prato2018}. Additionally, the C-MMNL model indicates that cyclists on utilitarian trips (regardless of the light conditions) to a larger extent prefer shorter routes compared to cyclists on leisure (but non-loop) trips. Protected bicycle tracks are the most favoured type of bicycle infrastructure, and the preference is even more pronounced for utilitarian trips (and even more so if a utilitarian trip takes place in the darkness). The light conditions alone hardly influence the preference for this attribute. The most avoided surface are cobblestones, but on utilitarian trips in the dark, the gravel surface is just as much of a deterring factor as cobblestones on non-utilitarian trips during the day. As for the land-use, areas near water are preferred over the high urban areas, but less so during the dark hours. Finally, cyclists indicate a heterogeneous behaviour for routes going through low-rise urban areas, with the context-dependent differences being less pronounced than for the other network attributes of interest.

Applying the proposed C-MMNL model enables behavioural analysis of many scenarios, without a significant increase in the computational burden of the estimation (2,329 seconds for MMNL vs. 2,792 seconds for C-MMNL), indicating that even though the C-MMNL model allows for flexible interactions between attributes, there is hardly an increase in the complexity of the model. Once the model has learned the structure from the input data, an analysis of different scenarios for context attributes (and their combinations) is possible without the necessity to reestimate the model.

\begin{figure}[H]
    \centering
        \includegraphics[width=0.6\textwidth]{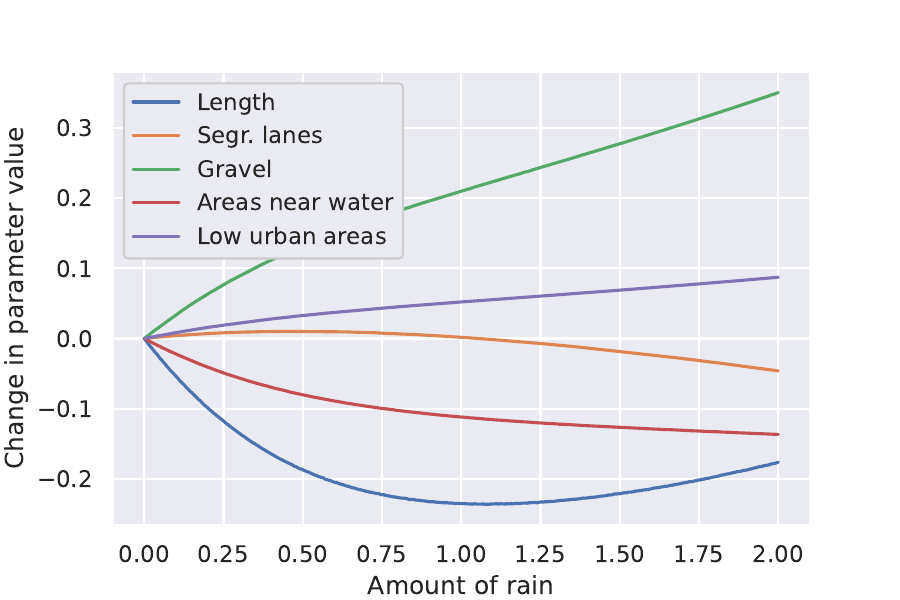}
    \caption{Change in the preference parameters for chosen model attributes, depending on the value of the contextual variable ‘‘amount of rain’’. Due to data scarcity and for illustration purposes, the plot domain was limited to the interval $\left[0,2\right]$.} 
    \label{fig:Rain}
\end{figure}

\paragraph{Amount of rain}
As mentioned earlier, the scenarios presented in the results of the C-MMNL model in Table \ref{tab:model} represent the average marginal effect of the third context attribute "amount of rain". This variable was included in the estimation as a continuous variable, and its influence can be analysed for all of the model attributes separately. Figure \ref{fig:Rain} shows the absolute change in the preference parameters for network attributes specified for interaction with the context attributes, compared to the "zero" scenario (value of the variable "amount of rain" equal to 0). The plots show the average marginal effects (weighted average over all possible scenarios for the other two binary context attributes). The plots indicate that route distance is highly affected by just some amount of rain and even more so in areas near water, which are likely more exposed to weather conditions. On the contrary, routes with a gravel surface (mostly present in areas with greenery), and in low urban areas to some extent, counteract the negative effect of route distance, maybe because these areas provide a better shield to weather conditions.

%\subsubsection{Trip context}
%Ton et al.~\textcite{Version2017} list possibly relevant contextual trip attributes for the case of route choice behaviour of cyclists, such as weather, daylight, trip purpose, or cycling season. In our model, we reveal the trip context information from the GPS data and we consider the following trip context attributes: precipitation, darkness and peak hours. %All the attribute values were computed at the first point of a trip and translated to binary variables.

%Table~\ref{tab:my_label} shows preliminary results obtained on a subsample of 1000 users with less than 10 choice situations. Early experiments with the full dataset indicate that, while MSL estimation takes several days (Biogeme), our SVI algorithm (PyTorch) run is less than 1h and achieves comparable results. Analysing the context-dependent shifts $\bs\mu_t$ from C-MMNL shows that precipitation decreases the preference parameter associate with length from -2.876 to -8.245, while darkness completely switches the preference for routes containing green (and usually poorly lit) areas from a positive coefficient (11.395) to a very negative one (-113.622). 

%\section{Model validation}
%No need for a separate section - part of case studies?
%\section{Discussion}
%\footnotes{\tr This should explore the significance of the results of the work, not repeat them. A combined Results and Discussion section is often appropriate. Avoid extensive citations and discussion of published literature.}
\section{Conclusion}
\label{sec:conclusion}
This paper presented an effective way to include contextual effects in the model while retaining the structure and properties of the traditional mixed multinomial logit model. The proposed context-aware Bayesian mixed multinomial logit model uses a neural network to systematically map the context corresponding to the choice occasion to the expected value of the context-dependent shifts on the base parameters. The model allows for complex interactions between attributes of different types while the results are delivered in an easily interpretable form (a model feature pointed out as important, \cite{han2022neural}). It learns the underlying structure in the input data and the analysis of different scenarios for context variables is possible without an additional computational cost. This can save the modeller the time usually spent on testing multiple parameter interactions. The source code for estimating the C-MMNL model, including a few examples, was made publicly available.

We performed an extensive simulation study on two datasets. These examples illustrate that the proposed method requires only a simple utility specification and a set of context variables. The results are provided in an easily-interpretable structure, despite modelling complex non-linear patterns. Furthermore, we show that the proposed C-MMNL model is able to recover the nominal preference parameters, as well as the context-dependent shifts, for both discrete and continuous variables. It extrapolates well on context scenarios unknown to the model at the estimation stage. The C-MMNL model improves both the individual and general predictions on a hold-out sample over variations of the traditional MMNL.

We applied the proposed C-MMNL model on a large-scale revealed preference dataset of 119,448 bicycle trajectories from 8,555 individuals. By making use of Stochastic Variation Inference (SVI) and GPU-hardware acceleration, we are able to scale the proposed model to handle a dataset of this size and estimate a C-MMNL model with 14 parameters and three context attributes in 47 minutes. The results are highly interpretable and it is possible to analyse the results from different perspectives: as fully-defined isolated context scenarios, in the form of average marginal effects for some of the variables, as well as on a domain range for continuous variables.

Future work could encompass the application of the C-MMNL model in other fields, where the effect of context attributes is crucial to understand the individual behaviour. Furthermore, the assumption that all individuals shift their preference parameters in the same way when faced with a given choice context could also be relaxed, allowing for complex interactions between the individual characteristics and the context information.

\paragraph{Acknowledgments} The authors wish to acknowledge Laurent Cazor for his contribution to the early stage of the data generation process in the simulation study.

\newpage
\printbibliography[title=References]

\end{document}